\documentclass{article}
\usepackage{spconf,amsmath,graphicx}
\usepackage{subcaption}
\usepackage{booktabs}
\usepackage{times}
\usepackage{epsfig}
\usepackage{graphicx}
\usepackage{amsmath}
\usepackage{amssymb}
\usepackage{subcaption}
\usepackage{enumitem}
\usepackage{tablefootnote}
\usepackage{multirow}
\usepackage{capt-of}
\usepackage[table]{xcolor}% http://ctan.org/pkg/xcolor

\title{NBD-GAP: Non-Blind Image Deblurring Without Clean Target Images}
\name{Nithin Gopalakrishnan Nair, \hspace{0.1cm} Rajeev Yasarla\hspace{0.16cm} and \hspace{0.16cm} Vishal M. Patel \thanks{This work was supported by the NSF CAREER award 2045489.}}
\address{Dept. of Electrical and Computer Engineering, Johns Hopkins University, MD, USA \\\texttt{\{ngopala2, ryasarl1, vpatel36\}@jhu.edu}}
\begin{document}
 \ninept
\maketitle
\vspace{-0.5cm}
\begin{abstract}
%  With the recent interest in research on long-range surveillance and face reconstruction, the idea of utilizing deep networks to tackle this problem has gained significant interest.
In recent years, deep neural network-based restoration methods have achieved state-of-the-art results in various image deblurring tasks. However, one major drawback of deep learning-based deblurring networks is that large amounts of blurry-clean image pairs  are required for training to achieve good performance. Moreover, deep networks often fail to perform well when the blurry images and the blur kernels during  testing  are very different from the ones used during training. This happens mainly because of the overfitting of the network parameters on the training data. In this work, we present a method that addresses these issues.
 We view the non-blind image deblurring problem as a denoising problem. To do so, we perform Wiener filtering  on a pair of blurry images with the corresponding blur kernels. This results in a pair of images with colored noise.  Hence, the deblurring problem is translated into a denoising problem. We then solve the denoising problem without using explicit clean target images. Extensive  experiments are conducted to show that our method achieves results that are on par to the state-of-the-art non-blind deblurring works. 
%This paper aims to aid upcoming research in atmospheric turbulence mitigation and help researchers choose the suitable data generation method for training deep learning models. The codes for the simulation methods, source codes for the networks and the pre-trained models will be publicly made available at:
% But most of these algorithms are computationally expensive and are unfeasible for creating datasets to train deep networks. This paper focuses on some recent works on modeling turbulence distortion for facial images and analyzing their performance in real-world face verification datasets. Specifically, we train state of the art networks using synthetic images generated using different turbulence modelling techniques and test the performance of the trained networks on real world data..
% We also consider images captured at different ranges and analyze the performance of networks trained using different simulation algorithms.
\end{abstract}
\begin{keywords}
Non blind deblurring, Wiener Deconvolution, No-reference.
% \vspace{-4mm}
\end{keywords}
\section{Introduction}
\label{sec:intro}

Motion blur is a common and prominent problem that occurs in hand-held photography. It destroys the aesthetics of the image and adversely affects the performance of many computer vision applications \cite{kupyn2018deblurgan,vasiljevic2016examining}. In this work, we focus on the case of uniform blur where the blurred image is represented by,
 \setlength{\belowdisplayskip}{0pt} \setlength{\belowdisplayshortskip}{0pt}
\setlength{\abovedisplayskip}{0pt} \setlength{\abovedisplayshortskip}{0pt}
\begin{equation}\label{eq:blur}
    y =k*x + n,
\end{equation}
where $y$ is the blurry image, $x$ is the latent clean image, $k$ is the blur kernel, $n$ is the additive white Gaussian noise, and  $*$ denotes the convolution operation.  Existing deblurring methods that restore the clean image $x$ can be grouped into blind and non-blind deblurring methods.  Non-blind deblurring methods take both the blurry image $y$ and the blur kernel $k$ as input to restore $x$.  On the other hand, blind deblurring is a more difficult problem that requires only the blurry image $y$ as input to restore $x$.  In this paper, we mainly focus on the non-blind deblurring problem. 

%Non-blind deblurring is an ill-posed problem, since the additive noise is random which makes it difficult to solve.
  \begin{figure}[t]
    \centering
    \includegraphics[width=\linewidth]{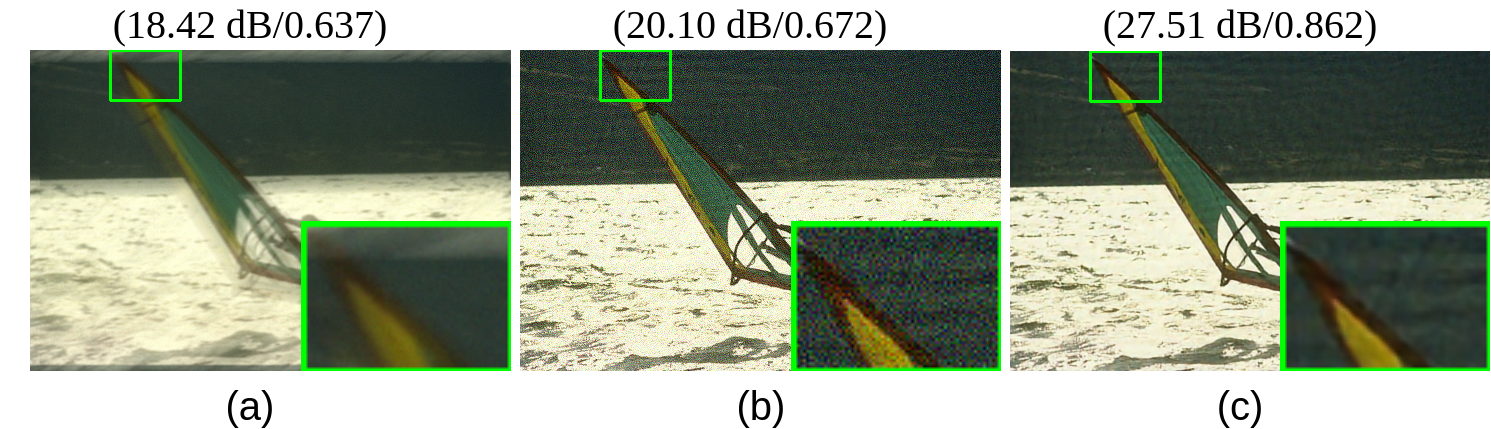}
        \vspace{-0.5cm}
    \caption{Sample output from our restoration network. (a) Blurry  input image.  (b) Wiener filtered image. Note the artifacts and noise after Wiener filtering. (c) Restored Image. }
    \label{fig:combined}
    \vspace{-0.5cm}
\end{figure}

Early non-blind deblurring algorithms used statistical techniques to derive minimum mean square error (MMSE) solutions such as the Wiener filter \cite{Wiener1949extrapolation} and Richardson-Lucy algorithm \cite{richardson1972bayesian}. Although these algorithms work well for very low noise levels, they add a significant amount of colored noise when the noise level increases. Also, these methods suffer from serious ringing artifacts and cannot deal with degradations caused by large motions. Another line of deconvolution algorithms focus on developing effective image priors \cite{krishnan2009fast,buades2005non,roth2005fields,sun2014good} using natural image statistics. But, these priors are very reliant on the distribution of the natural images and often lead to highly non-convex optimization problems, hence requiring expensive computational power. %to obtain good results. 

Recently neural networks have been used for various image deblurring tasks \cite{zhang2017learning,mohan2021deep,dong2021deep} and have achieved state-of-the-art results in different deblurring scenarios \cite{argaw2021motion,dong2021learning,mohan2021deep}. There are two major techniques by which deep learning has been adopted for non-blind deblurring. One line of works develop a model which uses both the blurred image and the kernel as input to address the deblurring problem \cite{xu2014deep,kruse2017learning}. Another line of works decomposes the problem into a denoising problem using a deconvolution algorithm or deep neural network.  Then address the deconvolution and denoising problems separately  \cite{schuler2013machine,zhang2017learning,son2017fast}. These techniques require large amounts of paired data, i.e., blurred images and their corresponding clean target pairs, to train the deep networks. These algorithms also fail to generalize well when the blur kernels during testing differ from the training kernels. Non-blind deblurring methods are often used as a black-box module for solving the blind deblurring algorithms \cite{lai2016comparative,cho2009fast,vasu2018non,fergus2006removing}
where the kernels and the latent images are found iteratively until the latent clean image is restored. Hence, there exists a need to develop non-blind deblurring algorithms that generalize over unseen images and kernels and learn with less data. To this end, we propose an approach that can be trained without clean target images and can generalize well to unseen data and kernels. 

%We proceed to describe in detail how we address both these problems in the coming sections.
% Hence we propose an approach [TODO: are we addressing this].}
% {\color{blue}\sout{Non-blind deblurring methods form the basis for solving blind deblurring cases also \cite{lai2016comparative,cho2009fast,vasu2018non,fergus2006removing}, where the kernels and latent images are found iteratively until the latent clean image is restored. Hence there exists a need to develop new non-blind deblurring techniques that could generalize well for different kernel basis.}}

\begin{figure*}[t!]
	\centering
		\includegraphics[width=.8\textwidth]{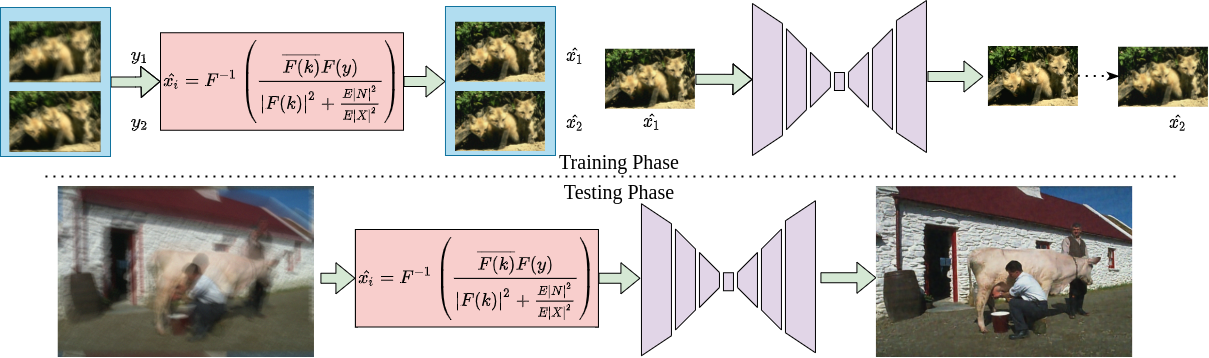}
	\centering
\vskip-8pt	\caption{An overview of the proposed network. During training, two input blurred images are processed through Wiener filtering to produce the corresponding two noisy images. Then these two noisy images are used as input and targetto train a network.  During  testing, for restoration, a single image is Wiener filtered and passed through the denoising network to obtain the restored image.}
	\label{fig:pipeline}
	\vskip -10pt
\end{figure*}
In this work, we follow the second line of techniques as mentioned above and propose a Non-blind image deblurring technique. Our method doesn't require the actual clean target data corresponding to the blurry input images. Specifically, we first perform  Wiener filtering to deblur two different blurry images of the same scene and get a pair of images with colored noise. Hence, we decompose the deblurring problem into a denoising problem. The key idea behind our denoising network is motivated by the work of Lehtinan et al. \cite{lehtinen2018noise2noise} where the authors have performed denoising of images corrupted by both Gaussian as well as colored noise without explicit training using clean target images. The main idea is that when a network is trained with multiple pairs of noisy images of the same latent scene, the expected loss between the network output and the noisy target is approximately equal to the expected loss between the network output and the corresponding clean target. The network hence learns the distribution of noise and can perform denoising. Inspired by this, we utilize Wiener deconvolution to produce a pair of images with zero centered colored noise. We then treat one of the outputs of Wiener deconvolution as source and the other as the target to train our network.

\section{Proposed method}
\subsection{Deconvolution as a denoising problem}
%In this paper, we use the mininum least square estimate and proceed to derive the solution as follows.  With the assumption that the blur is uniform, given a blurry image $y$, and it's corresponding blur kernel $k$, in non-blind deblurring problem, our objective is to retrieve clean image $x$, where they are related by the equation, $y = k*x + n$, where  $n$ is the additive white Gaussian noise.
%In the case of non blind deblurring information about the blur kernel $k$ is available. 
 Given $y$ and $k$ in \eqref{eq:blur}, our objective is to restore $x$.  A simple solution to this problem is finding an inverse filter $g$ where,
\begin{equation}
    \hat{x} =g*y,
\end{equation}
such that we can make an estimate of the clean latent image represented by $\hat{x}$. The ideal solution for $\hat{x}_{ideal}$ minimizes the expected mean square error between the clean latent image $x$ and the estimate
\begin{equation}
    \hat{x}_{ideal} = \min \hspace{1mm}E|x-\hat{x}|^2.
\end{equation}
% We convert the variables $x,k,y,g,n$ and $\hat{x}$ to frequency domain denoted by $X,K,Y,G,N$ and $\hat{X}$ respectively. the expected mean square error is then denoted by 
% \begin{equation}
%   \begin{array}{l}
%     E|X-\hat{X}|^2 =  E|X-GY|^2=E|X-G(KX+N)|^2 \\
%     \hspace{2cm} = E|(1-GK)X + GN|^2 
%     \end{array}
% \end{equation}
% Expanding the quadratic, and assuming that the signal and noise are independent from each other and optimizing with respect to G we get, 
% \begin{equation}
%  G = \frac{K^C}{|K|^2+\frac{|X|^2}{|N|^2}}
% \end{equation}
% Here C denotes the complex conjugate operator. This refers to the Wiener deconvolution process. Now applying the Wiener filter on the blurred image, we get
% \begin{equation}
%   \begin{array}{l}

% \hat{X}_{ideal}=GY =  \frac{K^C}{|K|^2+\frac{E|N|^2}{E|X|^2}}KX +  \frac{K^C}{|K|^2+\frac{E|N|^2}{E|X|^2}}N\\
%     \hspace{2cm} =X-\frac{\frac{E|N|^2}{E|X|^2}}{|K|^2+\frac{E|N|^2}{E|X|^2}}X + \frac{K^C}{|K|^2+\frac{E|N|^2}{E|X|^2}}N\\
%     \hspace{2cm} \approx X + \frac{K^C}{|K|^2+\frac{E|N|^2}{E|X|^2}}N
% \end{array}
% \end{equation}
% Since the term $\frac{E|N|^2}{E|X|^2}$ is negligible, it can be approximated to zero. Now convering back to signal domain we get,
\noindent This inverse filtering problem can be easily solved in the Fourier domain and the solution for the inverse filter $g$ is the Wiener filter defined by 
\begin{equation}
 G = \frac{K^H}{|K|^2+\frac{E|X|^2}{E|N|^2}},
\end{equation}
where $G,K,X,N$ denote the variables in the Fourier domain and $H$ denotes the Hermitian operator (detailed derivation is given in the supplementary document). The corresponding solution turns out to be  $\hat{x}_{ideal} = x + n_c,$ where 
\begin{equation}
n_c = IFFT\left(\frac{K^HN}{|K|^2+\frac{E|N|^2}{E|X|^2}}\right).
\label{eq:Wiener}
\end{equation}
Here, $n_c$ represents the inverse Fourier transform of the colored noise present along with the image after Wiener filtering. Now, since $\frac{E|N|^2}{E|X|^2}$ is an even function, the Fourier transform of $n_c$  is odd, and the inverse Fourier transform is also odd. Hence, its mean is zero. Thus the output of Wiener filtering contains the latent image corrupted with zero mean colored random noise. Hence by performing Wiener filtering, the deblurring problem converts into a denoising problem. One crucial aspect of Wiener filtering is estimating the noise to signal ratio (NSR) term defined by  $\frac{E|N|^2}{E|X|^2}$. In our work, similar to \cite{son2017fast}, we first find a median filtered estimate of the blurred image and compute its variance with the original blurred image to estimate NSR.  Multiple works have previously taken this approach to simplify the deconvolution problem to a denoising problem by utilizing various transformations such as shearlet  \cite{patel2009shearlet}, and wavelet  \cite{neelamani2004forward}. 

  Given blurry images $y$ with the corresponding blur kernels $k$, we first convert the deconvolution task into a denoising by applying Wiener filtering as follows,
% \vspace{-1em}
 \begin{equation}
     \hat{x}=F^{-1} \left( \frac{\overline{F(k)}F(y)}{|F(k)|^2+\frac{E|N|^2}{E|X|^2}}\right),
 \end{equation}
%  \vspace{-0.8em}
 where $F$ denotes the Fourier transform. In practical scenarios, the output of Wiener filtering is often corrupted with ringing artifacts along with the coloured noise. The colored noise could be removed to an extent by training the network using another set of noisy images as targets as the output like in the case of Lehtinan \textit{et al}\cite{lehtinen2018noise2noise}. 
 \vspace{-0.4em}

\begin{table*}[t]
   \centering
   \caption{Quantitative evaluation on  Levin et al\cite{levin2009understanding} dataset consisting of 32 images in terms of PSNR, SSIM. Please note that all the deep learning based methods used for comparison ad achieves better result than us in the $\sigma=12.75$ dataset is a supervised technique. MD-Motion deblurring, NBD- Non-Blind Deblurring. } 
 \resizebox{1\textwidth}{!}{
   \begin{tabular}{|c|c| c| c| c| c |c |c |c| c| c| }
      \toprule
    %   \multicolumn{2}{|c|}{\cellcolor[HTML]{FFFFFF}{s}} 
    
    \multirow{2}{*}{ Noise }&  \multirow{2}{*}{ Metrics }& Supervised-MD&\multicolumn{5}{|c|}{Supervised-NBD} & \multicolumn{2}{|c|}{No clean target training-NBD}\\
   & & Kupyn\cite{kupyn2019deblurgan}& Gong \cite{gong2018learning}& Son \cite{son2017fast}  & IRCNN\cite{zhang2017learning1} &  Zhang\cite{zhang2017learning} & Kruse\cite{kruse2017learning} & EPLL \cite{zoran2011learning} & OURS\\
      \midrule
       \multirow{2}{*}{\textit{$\sigma=2.55$}} & PSNR &24.52&32.32&31.56&32.94&32.87&33.68&33.72&\textbf{34.97}\\
       & SSIM &0.712&0.913&0.896&0.912&0.919&0.922&0.929 &\textbf{0.950}\\
      \midrule
       \multirow{2}{*}{\textit{$\sigma=7.65$}} &
       PSNR &23.97&29.22&28.95&30.51&29.55&29.80&29.05&\textbf{30.21}\\
       & SSIM&0.641&0.852&0.837&0.875&0.860&0.857&0.845 &\textbf{0.883}\\
      \midrule
         \multirow{2}{*}{\textit{$\sigma = 12.75$}} &
        PSNR &22.21&27.13&27.35&\textbf{27.92}&27.77&27.95&26.53&26.91\\
         & SSIM& 0.542&0.791&0.802&0.821&0.814&0.812&0.778&\textbf{0.822}\\

      \bottomrule
   \end{tabular}
   \label{table:psnr_levin}}
\end{table*}

\subsection{Network architecture}
The overall pipeline of our network is shown in Figure \ref{fig:pipeline}. During the training process, we create two different blurry images using different blur kernels and random noise signals. Then, each of these blurred images are  Wiener filtered to generate two images of the same latent scene with different random coloured noise. We use a simple U-net based architecture as our base network. We then use one of these images as the input and the other image as target output of the network.

 \begin{figure*}[ht!]
\centering
\begin{minipage}[c]{0.15\linewidth}
\begin{center}
\includegraphics[width=1.05\textwidth,height = 1.2\textwidth]{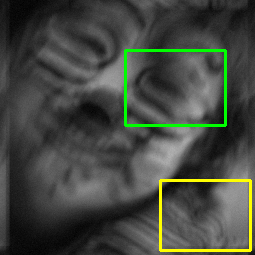}
\end{center}
\end{minipage}
\begin{minipage}[c]{0.84\linewidth}
\begin{center}
\includegraphics[width=0.16\textwidth,height=0.105\textwidth]{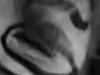}
\includegraphics[width=0.16\textwidth,height=0.105\textwidth]{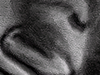}
\includegraphics[width=0.16\textwidth,height=0.105\textwidth]{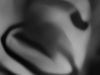}
\includegraphics[width=0.16\textwidth,height=0.105\textwidth]{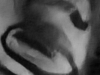}
\includegraphics[width=0.16\textwidth,height=0.105\textwidth]{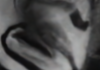}
\includegraphics[width=0.16\textwidth,height=0.105\textwidth]{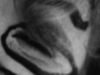}\\
\includegraphics[width=0.16\textwidth,height=0.105\textwidth]{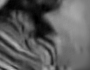}
\includegraphics[width=0.16\textwidth,height=0.105\textwidth]{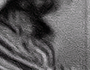}
\includegraphics[width=0.16\textwidth,height=0.105\textwidth]{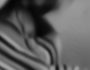}
\includegraphics[width=0.16\textwidth,height=0.105\textwidth]{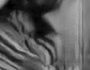}
\includegraphics[width=0.16\textwidth,height=0.105\textwidth]{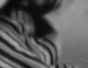}
\includegraphics[width=0.16\textwidth,height=0.105\textwidth]{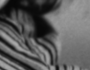}\\
\end{center}
\end{minipage}

\begin{minipage}[c]{0.15\linewidth}
\begin{center}
\includegraphics[width=1.05\textwidth,height = 1.2\textwidth]{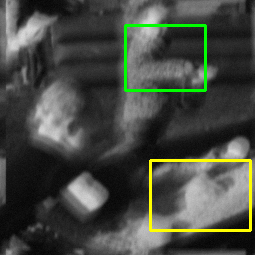}
\end{center}
\end{minipage}
\begin{minipage}[c]{0.84\linewidth}
\begin{center}
\includegraphics[width=0.16\textwidth,height=0.105\textwidth]{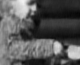}
\includegraphics[width=0.16\textwidth,height=0.105\textwidth]{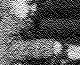}
\includegraphics[width=0.16\textwidth,height=0.105\textwidth]{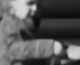}
\includegraphics[width=0.16\textwidth,height=0.105\textwidth]{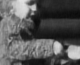}
\includegraphics[width=0.16\textwidth,height=0.105\textwidth]{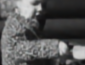}
\includegraphics[width=0.16\textwidth,height=0.105\textwidth]{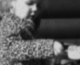}\\
\includegraphics[width=0.16\textwidth,height=0.105\textwidth]{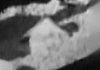}
\includegraphics[width=0.16\textwidth,height=0.105\textwidth]{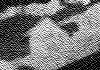}
\includegraphics[width=0.16\textwidth,height=0.105\textwidth]{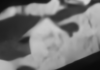}
\includegraphics[width=0.16\textwidth,height=0.105\textwidth]{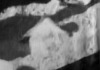}
\includegraphics[width=0.16\textwidth,height=0.105\textwidth]{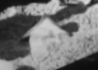}
\includegraphics[width=0.16\textwidth,height=0.105\textwidth]{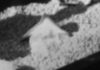}\\
\end{center}
\end{minipage}

\begin{flushleft}
\vskip-5pt

{ \hskip10pt Noisy Image \hskip35pt EPLL~\cite{zoran2011learning} \hskip30pt Son~\cite{son2017fast} \hskip30pt IRCNN~\cite{zhang2017learning1} \hskip25pt Kruse~\cite{kruse2017learning} \hskip30pt NBD-GAP(ours) \hskip15pt Ground-Truth}\\
\end{flushleft}
\vskip-10pt\caption{Qualitative evaluations on the Levin et al dataset \cite{levin2009understanding} for $\sigma=2.55$ noise level. }\label{fig:levin}
\end{figure*}

\subsection{Loss function}

Let $I_n$ and $\hat{I_2}$, denote two noisy images of dimensions $C\times H\times W$. The loss function used to train our network is
\begin{equation}
    \mathcal{L}_2 = \frac{1}{CHW}\sum_{i,j,k} \lVert I_1- \hat{I_2} \rVert^2.
\end{equation}
% Image gradient-based priors have been shown effective for multiple non-blind deblurring algorithms \cite{vasu2018non,gong2018learning}, and denoising works \cite{zoran2011learning}. Inspired from this and to ensure the sharpness of the predicted output, we use a modified version of the edge-preserving loss introduced in \cite{zhang2018densely} which is defined as 
%  \begin{equation}\label{tvloss}
% \mathcal{L}_{tv}= \sum_n \sum_{i,j} \lVert D_x(\hat{I})- D_x(I_n) \rVert_2 + \lVert D_y(\hat{I})- D_y(I_n) \rVert_2,
% \end{equation}
% where $D_x,D_y$ denote the horizontal and vertical derivatives of  the image.
\section{Experiments}
We use a batch size of 8 and use the AdamW optimizer with parameters $\beta_1=0.9$ and$\beta_2=0.999$. We set the initial learning rate to $10^{-4}$ and reduce it by half every 10 epochs. We train our network for a total of 60 epochs. ALl the experiments are run in 

\label{sec:dataset1}
\subsection{Training Dataset}
\label{sec:dataset}
For creating our training dataset, we follow the approach that has been followed in multiple non-blind deblurring works \cite{dong2021deep,zhang2017learning,zhang2017learning1,son2017fast}. We create a dataset of 1,000 images consisting of 600 images from the Waterloo exploration dataset \cite{ma2016waterloo} and 400 images from the Berkeley Segmentation dataset \cite{martin2001database}. We synthesize kernels of random sizes varying from $11 \times 11$ to $35 \times 35$ according to the method proposed by \cite{schmidt2015cascades} and convolve them with a patch of size $256 \times 256$ cropped from a clean image. We finally apply additive Gaussian noise of standard deviation $\{2.55,7.65,12.75\}$ to form blurry images. 

\subsection{Testing Dataset}
For testing our network, we use three different testing datasets. Similar to the approaches \cite{dong2021deep,zhang2017learning,son2017fast}, we use the popular benchmark datasets from Levin et al. \cite{levin2009understanding}. Levin et al. \cite{levin2007image} consists of 4 grayscale images respectively. We utilize the 8 standard kernels released by \cite{levin2009understanding} and blur these images and add noise of different variances in the range of  $\{2.55,7.65,12.75\}$ to create blurry images. We also create a synthetic dataset using the 100 test images of the BSD dataset \cite{martin2001database} and blur it with a random blur kernel  generated using \cite{schmidt2015cascades}. Note that the kernels used for the testing are different from the ones used for training.

\begin{table*}[t]
   \centering
   \caption{Quantitative evaluation on  500 blurry images generated from the BSD500 dataset \cite{martin2001database} in terms of PSNR, SSIM. Please note that all the deep learning-based methods used for comparison  achieve better results than us in the $\sigma=7.65$ noise case are supervised techniques. MD denotes Motion deblurring and NBD denotes Non-Blind Deblurring.} 
 \resizebox{1\textwidth}{!}{
   \begin{tabular}{|c| c |c| c| c |c| c |c |c |c| }
      \toprule
    \multirow{2}{*}{ Noise }&  \multirow{2}{*}{ Metrics }& Supervised-MD&\multicolumn{5}{|c|}{Supervised-NBD} & \multicolumn{2}{|c|}{Self-supervised-NBD}\\
   & & Kupyn\cite{kupyn2019deblurgan}& Gong \cite{gong2018learning}& Son \cite{son2017fast}  & IRCNN\cite{zhang2017learning1} &  Zhang\cite{zhang2017learning} & Kruse\cite{kruse2017learning} & EPLL \cite{zoran2011learning} & OURS\\
      \midrule
           \multirow{2}{*}{\textit{$\sigma=2.55$}}&PSNR &23.25&27.61&27.40&27.85&27.55&27.22&27.49&\textbf{30.27}\\
       &SSIM&0.712&0.841 &0.831&0.854&0.840&0.825&0.839&\textbf{0.862}\\
      \midrule
           \multirow{2}{*}{\textit{$\sigma=7.65$}}&PSNR &23.97&25.73&25.51&25.63&26.70&25.61&25.62&\textbf{26.81}\\
       &SSIM&0.641&0.751 &0.729&0.758&\textbf{0.802}&0.738&0.746&0.761\\
      \midrule
            \multirow{2}{*}{\textit{$\sigma=12.75$}}&PSNR &22.21&24.35&24.42&24.66&24.60&24.53&24.71&\textbf{24.82}\\
        &SSIM&0.542&0.649&0.641&0.660&0.652&0.677&\textbf{0.688}&0.685\\

      \bottomrule
   \end{tabular}
   \label{table:psnr_bsd}}
\end{table*}
  \begin{figure*}[ht!]
\vskip-5pt
\begin{minipage}[c]{0.1625\linewidth}
\begin{center}
\includegraphics[width=1.02\textwidth,height = 1\textwidth]{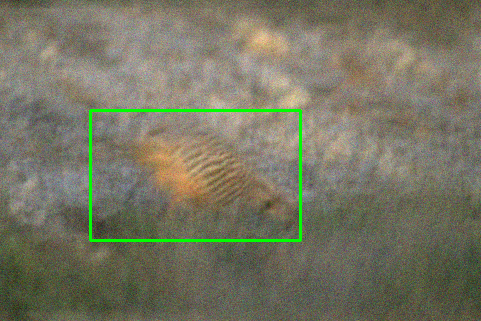}
\end{center}
\end{minipage}
\begin{minipage}[c]{0.8025\linewidth}
\begin{center}
\includegraphics[width=0.16\textwidth,height = 0.096\linewidth]{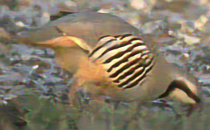}
\includegraphics[width=0.16\textwidth,height = 0.096\linewidth]{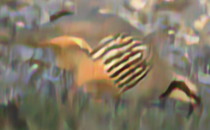}
\includegraphics[width=0.16\textwidth,height = 0.096\linewidth]{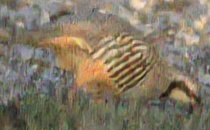}
\includegraphics[width=0.16\textwidth,height = 0.096\linewidth]{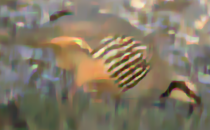}
\includegraphics[width=0.16\textwidth,height = 0.096\linewidth]{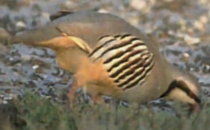}
\includegraphics[width=0.16\textwidth,height = 0.096\linewidth]{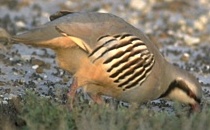}\\
\includegraphics[width=0.16\textwidth,height=0.096\textwidth]{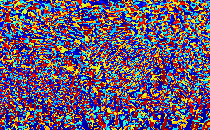}
\includegraphics[width=0.16\textwidth,height=0.096\textwidth]{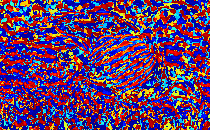}
\includegraphics[width=0.16\textwidth,height=0.096\textwidth]{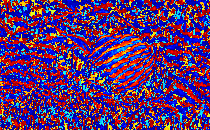}
\includegraphics[width=0.16\textwidth,height=0.096\textwidth]{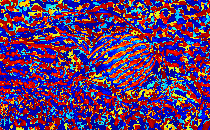}
\includegraphics[width=0.16\textwidth,height=0.096\textwidth]{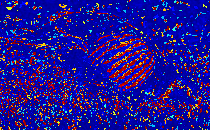}
\includegraphics[width=0.16\textwidth,height=0.096\textwidth]{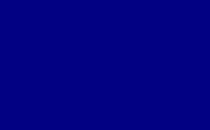}\\
\end{center}
\end{minipage}

\begin{minipage}[c]{0.1625\linewidth}
\begin{center}
\includegraphics[width=1.02\textwidth,height = 1.05\textwidth]{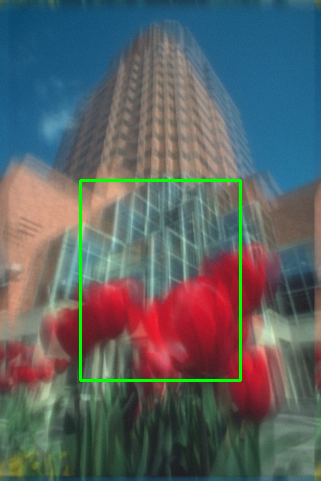}
\end{center}
\end{minipage}
\begin{minipage}[c]{0.8025\linewidth}
\begin{center}
\includegraphics[width=0.16\textwidth,height = 0.105\linewidth]{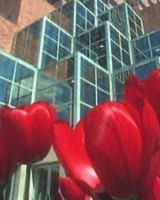}
\includegraphics[width=0.16\textwidth,height = 0.105\linewidth]{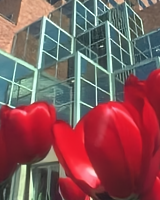}
\includegraphics[width=0.16\textwidth,height = 0.105\linewidth]{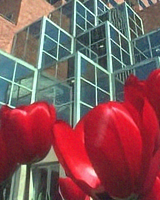}
\includegraphics[width=0.16\textwidth,height = 0.105\linewidth]{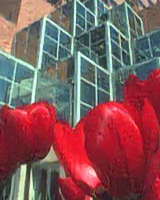}
\includegraphics[width=0.16\textwidth,height = 0.105\linewidth]{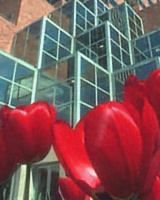}
\includegraphics[width=0.16\textwidth,height = 0.105\linewidth]{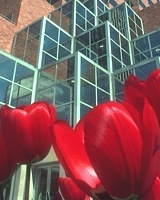}\\
\includegraphics[width=0.16\textwidth,height = 0.105\linewidth]{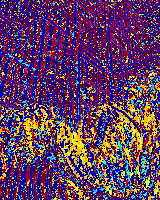}
\includegraphics[width=0.16\textwidth,height = 0.105\linewidth]{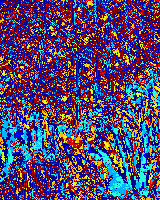}
\includegraphics[width=0.16\textwidth,,height = 0.105\linewidth]{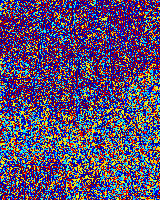}
\includegraphics[width=0.16\textwidth,height = 0.105\linewidth]{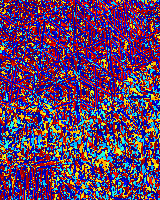}
\includegraphics[width=0.16\textwidth,height = 0.105\linewidth]{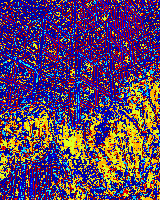}
\includegraphics[width=0.16\textwidth,height = 0.105\linewidth]{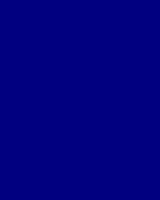}\\
\end{center}
\end{minipage}
\begin{minipage}[c]{0.025\linewidth}
\begin{center}
\includegraphics[width=\textwidth,height = 7\textwidth]{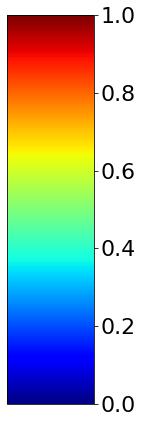}
\end{center}
\end{minipage}

\begin{minipage}[c]{0.1625\linewidth}
\begin{center}
\includegraphics[width=1.02\textwidth,height = 0.88\textwidth]{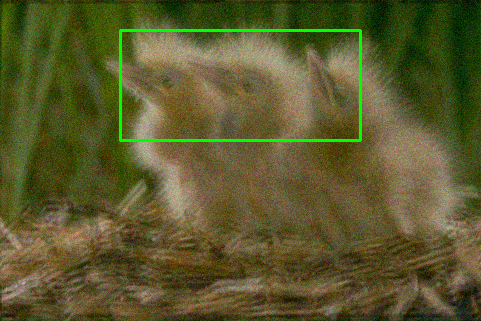}
\end{center}
\end{minipage}
\begin{minipage}[c]{0.8025\linewidth}
\begin{center}
\includegraphics[width=0.16\textwidth,height=0.086\textwidth]{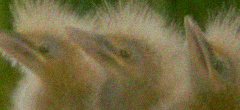}
\includegraphics[width=0.16\textwidth,height=0.086\textwidth]{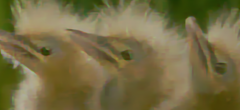}
\includegraphics[width=0.16\textwidth,height=0.086\textwidth]{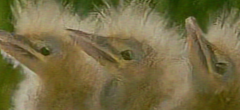}
\includegraphics[width=0.16\textwidth,height=0.086\textwidth]{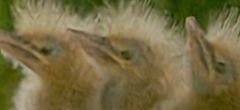}
\includegraphics[width=0.16\textwidth,height=0.086\textwidth]{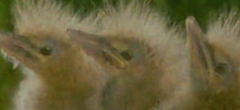}
\includegraphics[width=0.16\textwidth,height=0.086\textwidth]{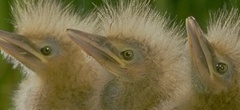}\\
\includegraphics[width=0.16\textwidth,height=0.086\textwidth]{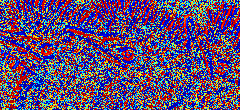}
\includegraphics[width=0.16\textwidth,height=0.086\textwidth]{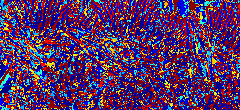}
\includegraphics[width=0.16\textwidth,height=0.086\textwidth]{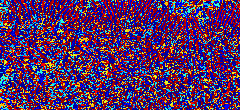}
\includegraphics[width=0.16\textwidth,height=0.086\textwidth]{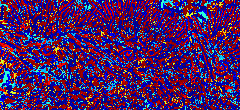}
\includegraphics[width=0.16\textwidth,height=0.086\textwidth]{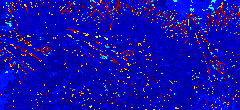}
\includegraphics[width=0.16\textwidth,height=0.086\textwidth]{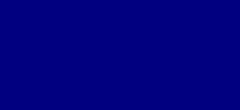}\\
\end{center}
\end{minipage}
  \begin{flushleft}
\vskip-5pt
{ \hskip10pt Noisy Image \hskip40pt EPLL~\cite{zoran2011learning}\hskip35pt Zhang~\cite{zhang2017learning1} \hskip25pt Son~\cite{son2017fast}  \hskip25pt Gong~\cite{gong2018learning} \hskip20pt NBD-GAP(ours) \hskip10pt Ground-Truth}\\
\end{flushleft}

\vskip-7pt\caption{Qualitative evaluations on the BSD500 dataset \cite{levin2009understanding} for $\sigma=7.65$ noise level and $\sigma=12.75$ noise level. The First two figures from top represent images of $\sigma=7.65$ noise level and the third figure has $\sigma=12.75$ noise level. The corresponding absolute error map from the ground truth is shown below each cropped patch. }\label{fig:bsd}
\end{figure*}

\subsection{Results}
  We evaluate the performance of our network by comparing qualitatively and quantitatively with different supervised works in non-blind deblurring. For comparison, we use a combination of classical non-blind deblurring algorithms \cite{zoran2011learning}, supervised deep networks designed for non-blind deblurring \cite{zhang2017learning,zhang2017learning1,son2017fast}, methods which utilize a combination of supervised deep learning and optimization algorithms \cite{gong2018learning} and one supervised motion deblurring network \cite{kupyn2019deblurgan}. Peak signal to noise ratio (PSNR) and structural similarity index measure (SSIM) are used to measure the performance of different methods quantitatively.

\subsubsection{Results on the Levin et al.\cite{levin2009understanding} dataset} 
 Table.\ref{table:psnr_levin} shows the quantitative results for different noise levels in the dataset prepared using images and kernels released by Levin et al. \cite{levin2009understanding}. 
 We can from Figure \ref{fig:levin} that conventional patch-based method EPLL \cite{zoran2011learning}  but creates a small amount of unwanted artifacts. Son et al. \cite{son2017fast} converts the deblurring problem to a denoising problem, similar to us. But we can see that some noise is still retained in the output. Note that the deep learning-based method IRCNN \cite{zhang2017learning1} achieves better results than us in the $\sigma=7.65$ noise case.  However, IRCNN is a supervised method and is trained with a lot of paired training data.

% \begin{table}[htp!]
% 	\caption{Quantitative evaluation on 640 blurred images generated from 80 clean images from SUN et al\cite{sun2013edge} dataset and 8 kernels from Levin et al.\cite{levin2009understanding} in terms of PSNR, SSIM .}
% 	\begin{center}
% 		\resizebox{1\linewidth}{!}{
% 			\begin{tabular}{|c|c c|c c|}
% 				\hline
% 				Noise level &\hspace{10mm}\textit{$\sigma=2.55$} &&\hspace{15mm}\textit{$\sigma = 7.65$}& \\
% 				\hline
% 				Metrics&PSNR&SSIM& PSNR&SSIM\\
% 				\hline
% 				Kupyn\cite{kupyn2019deblurgan}& 25.92&0.68&24.15& 0.48   \\
% 				 Gong \cite{gong2018learning}&31.50&0.867 &27.70 & 0.770\\
% 				% Lou \cite{lou2013video}& & &26.051 & 0.73\\
% 				Son \cite{son2017fast} &32.21& 0.883&28.45&0.767    \\
% 				IRCNN\cite{zhang2017learning1}&30.62&0.836 & \textbf{29.20}&\textbf{0.803} \\
% 				Zhang\cite{zhang2017learning}&32.09&0.882 &  28.95&0.784 \\
% 				Kruse\cite{kruse2017learning}&32.14&0.871 &  28.92&0.776 \\
% 				EPLL \cite{zoran2011learning} &32.24&0.879&28.55&0.769 \\
% 				Ours&\textbf{32.60} & \textbf{0.892}&28.62 &0.764 \\
% 				\hline
% 			\end{tabular}
% 		}
% 	\end{center}
% 	\label{table:sun}
% 	\vspace{-6mm}
% \end{table}

\subsubsection{Results on the BSD500 dataset \cite{martin2001database}} 
    Since both Levin et al.\cite{levin2009understanding} and Sun et al. \cite{sun2013edge} consist of gray scale images, to evaluate the performance on coloured images, we generate a test set using 100 test images from \cite{martin2001database} using kernels from \cite{schmidt2015cascades} for three different noise levels. As we can see in Table \ref{table:psnr_bsd} and Fig. \ref{fig:bsd}, our method is able to perform well and produce results with fewer artifacts and noise compared to \cite{son2017fast}.  Our method is even able to produce good results at higher levels of noise. The residual maps show that the absolute error is much lesser in our method when compared to the other techniques.

\section{Conclusion}
In this paper, we proposed the first work in non-blind deblurring, where we train a deep network without clean targets. We achieve this by converting the deblurring problem to a denoising problem using Wiener filtering.  We perform extensive experiments on multiple datasets to show the relevance of our network and show that our network can generalize well over multiple datasets.  Furthermore, it is shown that our network is able to produce good results even at higher levels of noise.
\vspace{-0.5cm}
% \clearpage
\bibliographystyle{IEEEbib}
{\footnotesize
\bibliography{egbib}
}
\end{document}